\documentclass[a4paper, 10pt, conference]{ieeeconf}      

\IEEEoverridecommandlockouts                              

\usepackage{graphics} 
\usepackage{epsfig} 
\usepackage{mathptmx} 
\usepackage{times} 
\usepackage{amsmath} 
\usepackage{amssymb}  
\usepackage{float}
\usepackage{tabularx}
\usepackage{booktabs}
\usepackage{multirow}
\usepackage{caption}
\usepackage{titlesec}
\usepackage{indentfirst}

\titlespacing*{\subsection}
{0pt}{1.0ex plus 1ex minus .2ex}{0.5ex plus .2ex}

\titlespacing*{\section}
{0pt}{1.0ex plus 1ex minus .2ex}{0.5ex plus .2ex}

\title{\LARGE \bf
Cooperative Grasping and Transportation using Multi-agent Reinforcement Learning with Ternary Force Representation
}

\author{Ing-Sheng Bernard-Tiong$^{*1}$, Yoshihisa Tsurumine$^{*1}$, Ryosuke Sota$^{1}$, Kazuki Shibata$^{1}$, and Takamitsu Matsubara$^{1}$
\thanks{$^{*}$Equal contribution.}
\thanks{$^{1}$IB, YT, RS, KS, and TM are with Department of Science and Technology, Graduate School of Science and Technology, Nara Institute of Science and Technology, Nara, Japan} 
}

\begin{document}

\maketitle
\thispagestyle{empty}
\pagestyle{empty}

\begin{abstract}
Cooperative grasping and transportation require effective coordination to complete the task. This study focuses on the approach leveraging force-sensing feedback, where robots use sensors to detect forces applied by others on an object to achieve coordination. Unlike explicit communication, it avoids delays and interruptions; however, force-sensing is highly sensitive and prone to interference from variations in grasping environment, such as changes in grasping force, grasping pose, object size and geometry, which can interfere with force signals, subsequently undermining coordination. We propose multi-agent reinforcement learning (MARL) with ternary force representation, a force representation that maintains consistent representation against variations in grasping environment. The simulation and real-world experiments demonstrate the robustness of the proposed method to changes in grasping force, object size and geometry as well as inherent sim2real gap. 
\end{abstract}

\section{INTRODUCTION}
Cooperative grasping and transportation involve multiple robots coordinating with each other to grasp and transport an object. This is crucial in scenarios where a single robot cannot independently manage heavy or bulky objects. 

Traditional cooperative transportation methods rely on model-based approaches that use planning or control theory to achieve coordinated transportation. One such approach involves leveraging force-sensing feedback \cite{leader_follower, force_implicit_2}, where robots use force sensors to perceive the applied force on an object by other robots to achieve coordination. While this method avoids the issues of delay or interruptions as in explicit communication, it requires pre-attachment, where robots are manually attached to the object, to stabilize force communication. On the other hand, in recent years, model-free methods based on MARL have been proposed. Previous work has demonstrated the use of MARL to accomplish cooperative grasping and transport without pre-attachment \cite{rl_dual_arm}; however, its reliance on explicit communication makes it susceptible to communication failures.

Achieving end-to-end cooperative grasping and transportation through force-sensing feedback is of great value because it avoids the reliance on explicit communication and manual pre-attachment. However, cooperative grasping and transportation based on force-sensing feedback is prone to interference from variations in grasping environment. This is because the force states representing the applied force between the robot's gripper fingers and the object are easily influenced by factors such as grasping pose, grasping force, and object size and geometry. Any subtle change in these factors causes significant changes in the grasping environment and the observed force states. Robots not exposed to such varied force observations are likely to fail the task.

\begin{figure}[tb]
    \centering
        \large
        \includegraphics[width=1.0\linewidth]{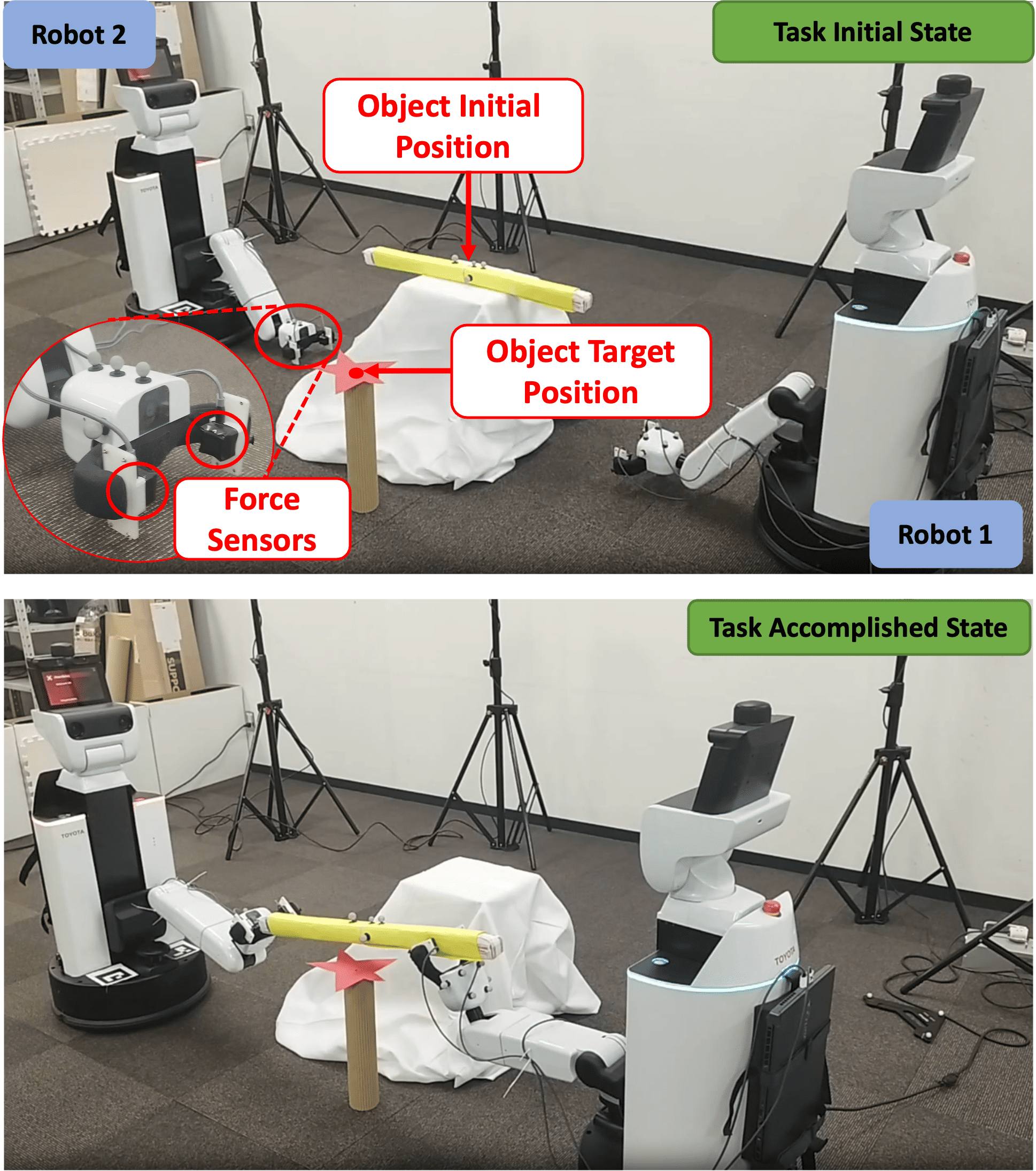}
        \caption{Cooperative grasping and transportation by two HSRs through implicit communication via force-sensing feedback.}
        \label{task_fig}
\end{figure}

In this study, we explore MARL for cooperative grasping and transportation tasks based on force-sensing with ternary force representation. The ternary data strictly restricted to discrete values \{-1, 0, 1\} can qualitatively represent the object motion that is required for cooperative transportation and ensure consistency in representation despite small variations. Herein, 1 and -1 represent motions in positive and negative directions respectively, and 0 indicates stationary motion. To resolve the partial observability introduced by the simplified ternary representation, we use MARL to train policy with rich force information during training and deploy them with only the ternary force representation during execution.

In our approach, the ternary force representation is derived from the raw signals of the force sensor located at gripper's fingers by applying ternary discretization to the delta force between consecutive time points \( t \) and \( t-1 \). We then employ asymmetric actor-critic architecture within centralized training decentralized execution (CTDE) MARL framework \cite{mappo}, where the actor uses ternary force representation, while the critic uses both the ternary force representation and the pre-discretization form of the ternary force representation, delta force, to mitigate partial observability. Since our method uses ternary representation during execution, robots can flexibly transport an object even if the grasping environment changes.

We confirm the effectiveness of our method through a cooperative grasping and transportation task in simulation and real-robot experiments using two Human Support Robots (HSR), as shown in Fig. \ref{task_fig}. The experimental results show that our method enables the robots to successfully grasp and transport objects despite variations in grasping environment.

Our contributions are as follows:
\begin{itemize}
  \item We propose a novel framework that is robust to variations in grasping environment for cooperative grasping and transportation tasks using force-sensing feedback. This method can be effectively applied even when there are changes in grasping force, geometry and size of object.  
  
  \item In the simulation experiment, we demonstrate the robustness of our method to variations in grasping force and object geometry.

  \item In the real-robot experiment, we demonstrate the robustness of our method to variations in object size and the inherent sim2real gap in grasping environment, including factors such as sensor models, object dynamics, and physical properties.
  
\end{itemize}

\section{RELATED WORK}
\subsection{Cooperative Object Transportation}
We categorize the research works into three categories, each of which is discussed in a separate paragraph:

\textit{1) Pushing Strategy Tasks:} It involves multiple robots working together to transport objects across the ground by exerting pushing forces. In this strategy, robots do not physically attach themselves to the object but instead apply force directly to its surface to move it. Previous work has explored controller design using hand-engineered models \cite{push_2} and MARL approaches \cite{shibata}. 

\textit{2) Pre-attachment Strategy Tasks:} It involves multiple robots distributed evenly to entrap the object with their end-effectors rigidly attached to the object. The object is transported by the resultant force generated from the coordinated pushing or pulling actions of each robot. The exemplary model is leader-follower \cite{leader_follower, leader_follower_3}. Robots periodically measure the force being exerted on the object, and apply force-updating law \cite{implicit_and_explicit} that uses only locally known terms to adjust the force being exerted by each individual robot. 
 
While the pre-attachment condition simplifies research by bypassing the need for grasping strategies, it significantly limits adaptability in real-world scenarios.



\textit{3) Grasping Strategy Tasks:}
It involves robots with arms and grippers working together to perform end-to-end cooperative grasping and transportation tasks. Since the robots start by grasping the object, they are not pre-attached, which provides greater flexibility in handling various object types, geometries, and sizes.

In the previous work \cite{rl_dual_arm}, MARL with observation sharing is leveraged to accomplish the task. However, the reliance on explicit communication renders the system vulnerable to delays and interruptions \cite{delay_1, delay_2}. Our task, while categorized under the grasping strategy, uses implicit communication through force-sensing feedback, eliminating the risk of communication failures.

\subsection{MARL for Cooperative Transportation}
MARL extends traditional reinforcement learning to scenarios involving multiple agents, making it an excellent approach for solving multi-robot cooperative tasks \cite{marl_overview, marl_survey}.


MARL has been applied to the pushing strategy task, where robots collaboratively push a payload to a target state \cite{shibata}. The framework uses Multi-agent Deep Deterministic Policy Gradient (MADDPG) with event-triggered communication and consensus-based control, allowing robots to execute tasks with varying agent numbers and minimal communication.

Parameter-sharing Deep Q Learning (DQN) is explored to tackle the pre-attachment strategy task \cite{push-and-pull}. Robots attached in a predefined lattice coordinate through "push-and-pull" forces, naturally generating implicit communication based on discrepancies between intended and actual actions.

The Dual-arm Deep Deterministic Policy Gradient (DADDPG) extends MADDPG to tackle the grasping strategy task, enabling independent agent training with sparse rewards, while sharing observations between agents \cite{rl_dual_arm}. This approach allows dual-arm robots to grasp and lift long rod-shaped objects without causing competition between the arms.

While previous studies rely on consensus-based explicit communication \cite{shibata} or observation sharing \cite{rl_dual_arm}, our method diverges significantly by leveraging a ternary force representation for implicit communication. This ternary representation ensures consistency across varying grasping conditions, making our method robust to changes in grasping force, geometry and size of object. By maintaining a consistent force representation, our method can be applied flexibly in environments where grasping conditions fluctuate.

\section{Method}
\subsection{Decentralized Partially Observable Markov Decision Processes (Dec-POMDP)}
Since each agent \(i\) (where \(i = 1, \dots, N\)) acts based on local observations in a multi-agent system, we model it using Dec-POMDP. It is defined by the tuple \(\langle N, S, A, O, R, P, \gamma \rangle\), where \(N\) is the number of agents, \(S\) denotes the global state of grasping environment, encompassing ground truth \(N\) agent grippers and object poses, forces in the environment, and \(o_i = O(s; i)\) is the local observation, which encompasses observed agent \(i\) gripper and object poses, force by sensors. The transition probability \(P(s' \mid s, A)\) represents the probability of moving from state \(s\) to next state \(s'\) given the joint action \(A = (a_1, \ldots, a_N)\). The reward function \(R = (r_1, \ldots, r_N)\) specifies rewards for each agent. \(\gamma\) is the discount factor. Each agent \(i\) aims to maximize its total expected return \(J_i(\theta_i) = \sum_{t=0}^{H} \gamma^t r_t^i\), where \(H\) is the time horizon.

\subsection{Proposed Framework}
\begin{figure}[tb]
    \centering
        \large
        \includegraphics[width=1.0\linewidth]{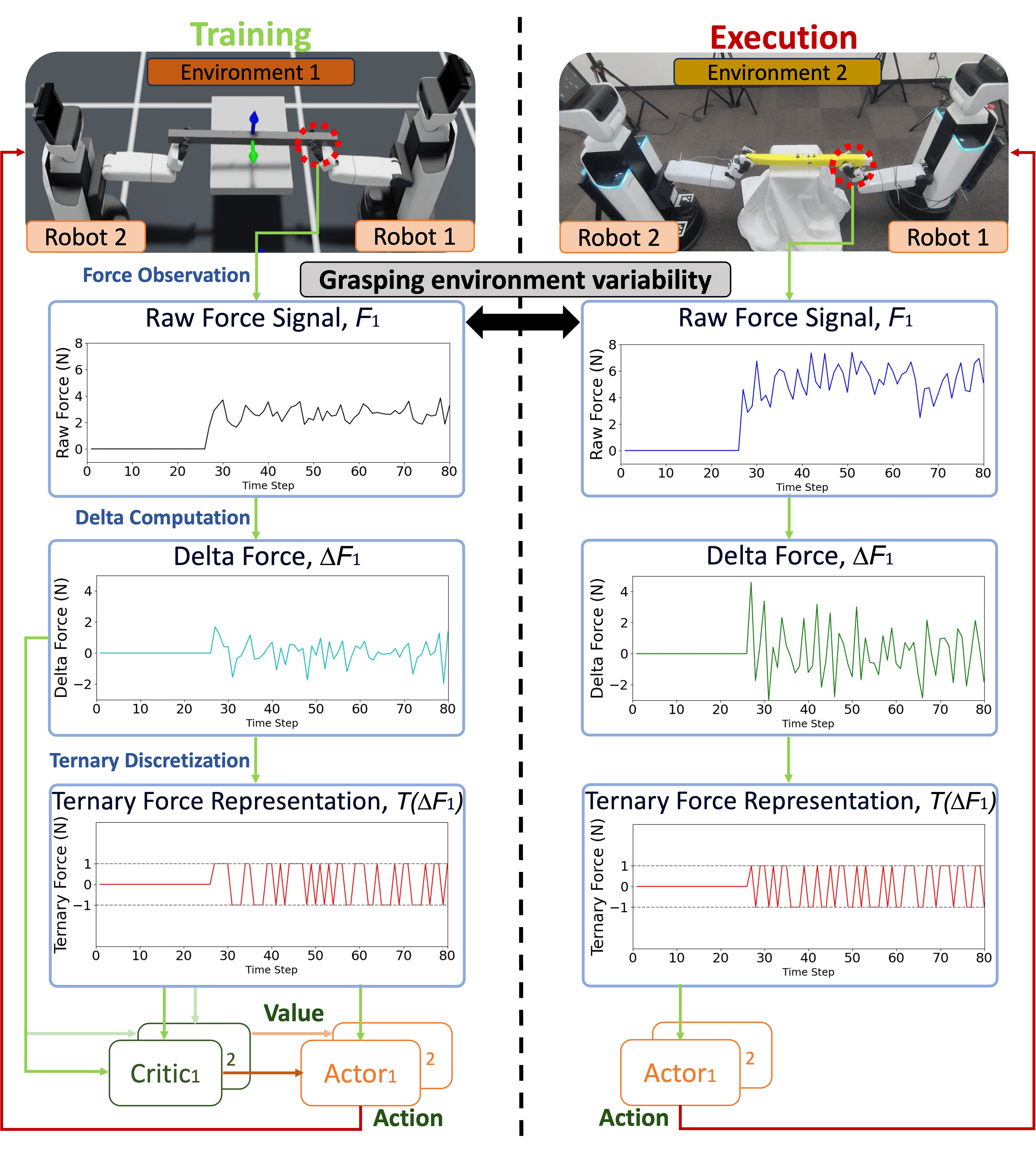}
        \caption{Asysmetric actor-critic-based MARL framework with ternary force representation. Even though the raw force signals differ between training and execution, converting raw force signals to the representation with consistent values of -1, 0, or 1 ensures consistency and retains informative values despite variations in the grasping environment.}
        \label{framework}
\end{figure}

We propose a novel reinforcement learning framework with asymmetric actor-critic architecture on top of MARL structure to exploit ternary force representation, as shown in Fig. \ref{framework}. In this framework, the asymmetric actor-critic approach is used to resolve the partial observability introduced by the simplified ternary representation. 

\subsubsection{Ternary Force Representation}
This representation is designed to provide a consistent representation of object motion information, which is crucial for effective coordination during transportation. It involves two key processing steps to transform raw force data into a consistent yet informative representation. The first step aims to capture essential motion information, the second step aims to discretize the motion information into ternary data to ensure consistency in representation despite small variations. The process is as follows:

\textit{Delta Computation:} From the streaming force sensor data of robot \(i\), we compute the delta force, \(\Delta F_i(t)\), between time \(t\) and time \(t-1\). This step captures the change in force, which partially encodes information about the motion of the object.

\textit{Ternary Discretization:} The delta force is then discretized into ternary force representation, \(T(\Delta F_i(t))\), to indicate the direction and presence of force changes. If the delta force is positive, it is represented by 1; if it is negative, by -1; and if there is no change, by 0. This ternary force representation serves as force observation, replacing raw force signals.

\subsubsection{Asymmetric Actor-critic with Ternary Force Representation}
Since the ternary force representation captures only directional information, the absence of magnitude data makes it challenging for robots to coordinate their actions with each other during object transportation. Consequently, this partial observability can affect the learning process, making it difficult to converge to a stable, effective policy.

To address the issue, we use the concept of asymmetric actor-critic \cite{AAC}, where the critic uses both direction information (ternary force representation) and magnitude information (delta force), while the actor only uses direction information. This approach allows robots to learn adjusting their action based on rich force information during training. During execution, they can coordinate with each other using only the ternary force representation.


\subsection{Reward Design}
The tasks of cooperative grasping and transportation can be decomposed into a series of sub-tasks: reaching, grasping, lifting, and transporting. Naturally, the reward structure can be bifurcated into individual and team-based rewards:
\begin{equation}
\begin{aligned}
r_i &= \underbrace{w_1 \cdot r_{\text{reach}} + w_2 \cdot r_{\text{grasp}}}_{\text{Individual reward}} + \\
    &\underbrace{w_3 \cdot r_{\text{grasp\_team}} + w_4 \cdot r_{\text{lift}} + w_5 \cdot r_{\text{pos}} + w_6 \cdot r_{\text{ori}}}_{\text{Team reward}}
\end{aligned}
\end{equation}

\noindent
where \(r_{\text{reach}}\) is the distance reward between end-effector and grasping point, \(r_{\text{grasp}}\) is the individual grasp reward, \(r_{\text{grasp\_team}}\) is the team grasp reward, \(r_{\text{lift}}\) is awarded when the object is successfully lifted a certain distance above the table, \(r_{\text{pos}}\) is the distance reward between object current position and target position, \(r_{\text{ori}}\) is the penalty reward to keep the object level during transportation, and $w_i$ ($i=1,\cdots,6$) is the corresponding weights. The team rewards can only be received if all robots grasp the object.

\subsection{Policy Optimization}
We employ the Multi-Agent Proximal Policy Optimization (MAPPO) algorithm \cite{mappo} to learn the policy. This approach follows the CTDE paradigm. The critic network is trained using the observations of all robots, which enables it to effectively evaluate the global state of the environment. Meanwhile, the actor network learns based on local observations, allowing each robot to make decisions independently. In this study, each agent has its own critic and actor networks.

\section{Simulation Experiment}
\subsection{Experiment Overview}
The experiment has two objectives. The first objective is to evaluate the effectiveness of the proposed ternary force representation and asymmetric actor-critic based on task performance. The second objective is to evaluate the policy’s robustness against grasping environment variations with different grasping forces and object geometries.

\begin{figure}[tb]
    \centering
        \large
        \includegraphics[width=1.0\linewidth]{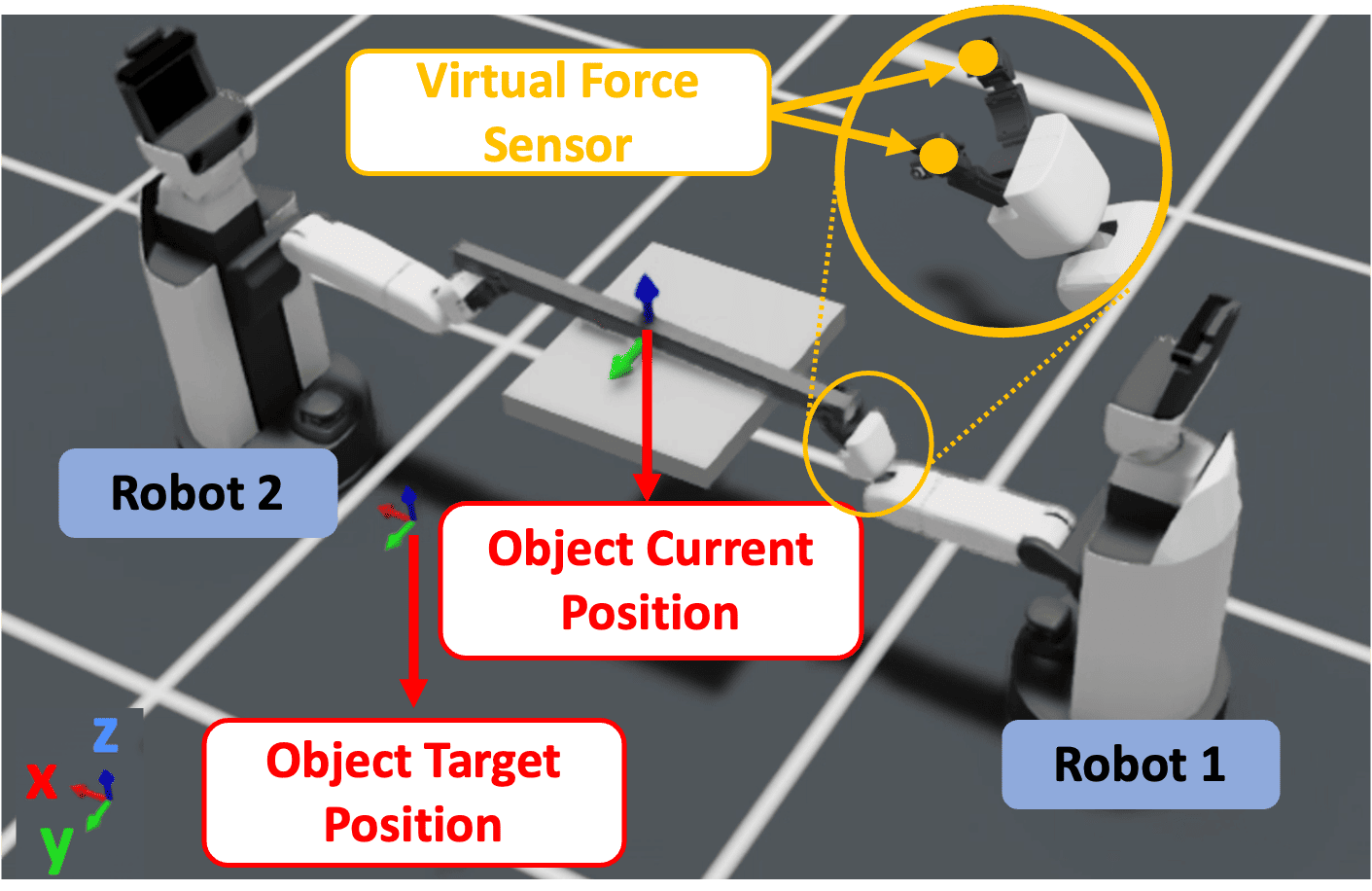}
        \caption{Simulation environment for cooperative grasping and transportation.}
        \label{sim_task_fig}
\end{figure}

\subsection{Experiment setup}
The task is to have two HSRs cooperatively reach, grasp, lift, and transport an object to a predefined target position in a 3D space. NVIDIA’s physics simulator, Isaac Sim, is utilized to construct a parallel simulation environment for policy training, as illustrated in Fig. \ref{sim_task_fig}.

Each robot observation, \(o_i\in \mathbb{R}^{27} \), includes the robot's arm joint position, gripper grasping status, robot base velocities, object grasping point position, object centroid position and orientation, object target position, force signals (without randomization) from a 3-axis force sensor on each gripper finger. Each robot action, \(a_i\in \mathbb{R}^{8} \), includes the robot's base and arm joint velocities, and gripper torque. 

The object is initially placed on the table. The object target position is uniformly distributed within the ranges of [-20, 20] (x-axis), [-80, 80] (y-axis), and [45, 80] (z-axis) relative to the object's initial position, as shown in Fig. \ref{sim_task_fig}. All measurements are in centimeters. The value for \(w_1\), \(w_2\), \(w_3\), \(w_4\), \(w_5\), \(w_6\) is 3.0, 4.0, 7.5, 9.5, 20 and 3.0, respectively. Empirically, higher coefficients for team rewards over individual rewards improve policy optimization.

We introduce the success rate to quantitatively evaluate performance, considering a task successful if the Euclidean distance between the object's position and target is under 6 cm, approximately the diagonal of the object's thickness.

\subsection{Baseline}
To evaluate the effectiveness of the proposed ternary force representation and asymmetric actor-critic, we compare the following baselines alongside our method:
\begin{itemize}
  \item Raw Force: MAPPO trained with raw force signals.
  \item No Force: MAPPO trained without force signals.
  \item Ternary Force: MAPPO trained with ternary force representation only.
  \item Ours: MAPPO with critic trained with delta and ternary force representation, and actor trained with ternary force representation.
\end{itemize}



\subsection{Result}
\subsubsection{Task Performance}
\begin{figure}[tb]
    \centering
        \large
        \includegraphics[width=1.0\linewidth]{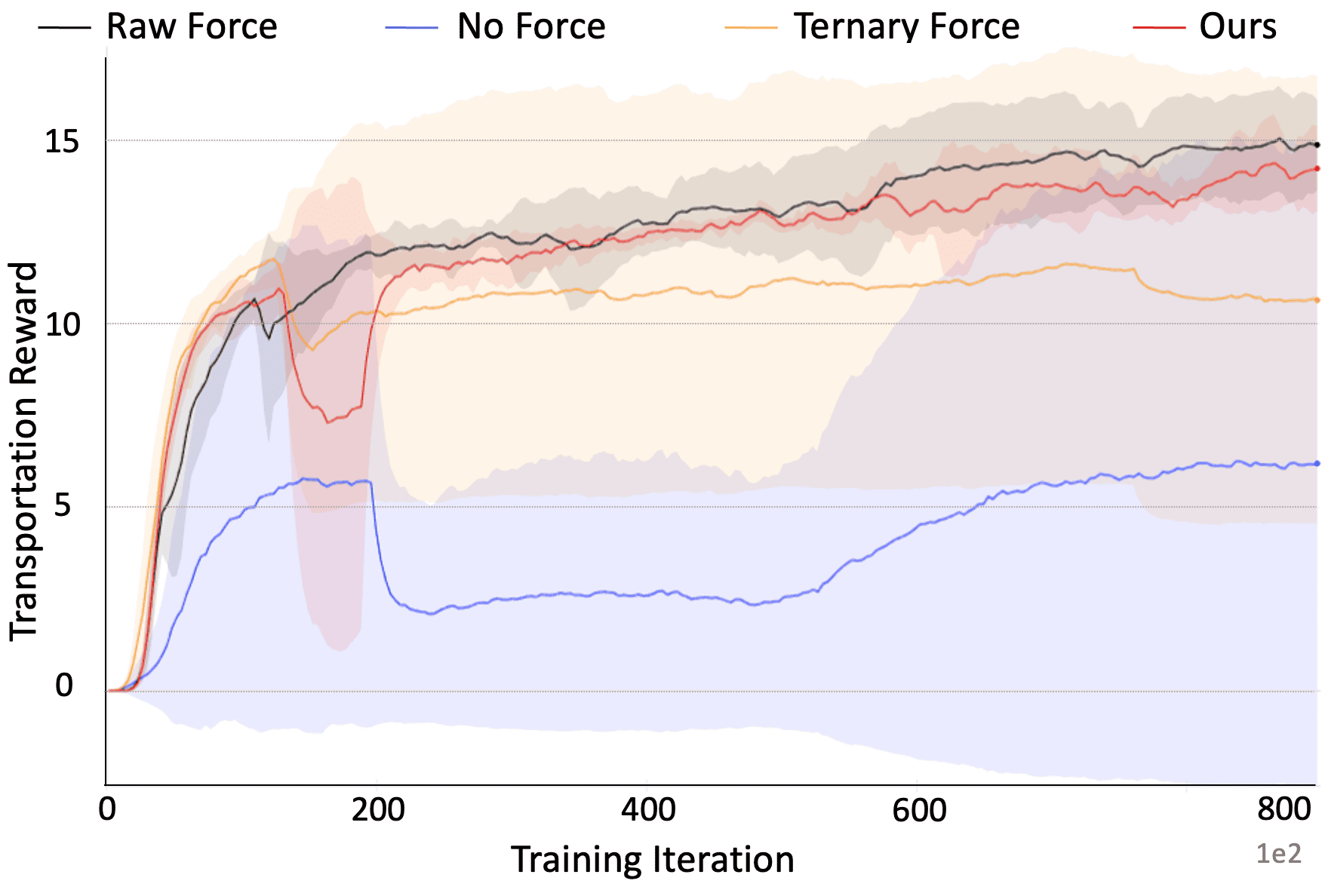}
        \caption{Cumulative rewards of evaluated methods.}
        \label{learning_curve}
\end{figure}

The transportation reward for our method and three comparison methods is shown in Fig. \ref{learning_curve}. For all methods, we conducted five training runs with the same setup but different seeds and presented the mean and variance across these runs.

To evaluate the task performance quantitatively, we measured the position error based on the Euclidean distance between the object's final position and the object's target position, as shown in Table \ref{position_error}. We assessed each method by averaging the position errors across 100 test episodes.

\begin{table}[!tp]
\centering
\caption{Object position error of evaluated methods.}
\label{position_error}
\begin{tabularx}{\columnwidth}{>{\centering\arraybackslash}X >{\centering\arraybackslash}c >{\centering\arraybackslash}c}
\toprule
\normalsize \textbf{Method} & \normalsize \textbf{Mean (cm)} & \normalsize \textbf{Variance (cm)} \\ 
\midrule
\textbf{Ours} & 5.21 & 0.32 \\ 
Raw Force & \textbf{4.55} & \textbf{0.21} \\ 
Ternary Force & 55.78 & 7.57 \\ 
No Force & 95.62 & 15.53 \\ 
\bottomrule
\end{tabularx}
\end{table}

The Raw Force's performance is the best because there was no partial observability caused by information loss. Our method's performance is comparable to the Raw Force, despite the actors only using ternary information. For the Ternary Force, robots could transport the object for a certain distance, but due to uncoordinated actions between robots, the object often fell during transportation. This demonstrates that feeding the delta force signals to the critic network during training helps address partial observability. For the No Force, although the robots were able to grasp the object, they could not coordinate with each other to transport it due to the absence of an implicit communication medium. 

\subsubsection{Evaluation of Policy's Robustness to Grasping Environment Variations}
To simulate the grasping environment variations, we altered the grasping environment and tested the trained policy under these altered environments. The variations included changes in grasping force and object geometry. We compared our method against the Raw Force because it represents the most direct use of force information. We evaluated each method by averaging the robot's object transportation success rate across 100 test episodes.

\paragraph{Variations in Grasping Force} 
Grasping force discrepancies are a common issue in robotic manipulation, where variations in the applied force can affect force signals, thus undermining coordination. 

To simulate such variations in grasping force, we scaled the grasping force during execution with scaling factors of 0.5, and 2, representing half and double the original grasping force, respectively. The result shown in Table \ref{grasping_force} indicates that in contrast to the Raw Force method, our method maintains high performance despite changes in grasping force.

\begin{table}[!tp]
\centering
\caption{Success rates of transportation with different grasping forces.}
\label{grasping_force}
\begin{tabularx}{\columnwidth}{>{\centering\arraybackslash}p{3.5cm} >{\centering\arraybackslash}X >{\centering\arraybackslash}X}
\toprule
\textbf{\normalsize Grasping Force Scale} &  \textbf{\normalsize Ours} & \textbf{\normalsize Raw Force} \\ 
\midrule
1 & \textbf{86\%} & \textbf{89\%} \\ 
0.5 & \textbf{85\%} & 40\% \\ 
2 & \textbf{80\%} & 30\% \\ 
\bottomrule
\end{tabularx}
\end{table}

\begin{table}[!tp]
\centering
\caption{Comparison of mean and variance of force observations with different grasping forces.}
\label{grasping_force_mean_var}
\begin{tabularx}{\columnwidth}{>{\centering\arraybackslash}p{0.8cm} >{\centering\arraybackslash}X >{\centering\arraybackslash}p{1.5cm} >{\centering\arraybackslash}X >{\centering\arraybackslash}p{1.5cm}}
\toprule
\multirow{2}{=}{\centering \scriptsize \textbf{\shortstack{Grasping\\Force\\Scale}}} & \multicolumn{2}{c}{\normalsize \textbf{Mean}} & \multicolumn{2}{c}{\normalsize \textbf{Variance}} \\ 
\cmidrule(lr){2-3} \cmidrule(lr){4-5}
 & \scriptsize \textbf{Ours} & \scriptsize \textbf{Raw Force} & \scriptsize \textbf{Ours} & \scriptsize \textbf{Raw Force} \\ 
\midrule
1 & \textbf{-0.015} & 2.111 & \textbf{0.815} & 1.311 \\ 
2 & \textbf{0.057} & 4.171 & \textbf{0.792} & 5.742 \\ 
\bottomrule
\end{tabularx}
\end{table}

We further investigate the effect of different grasping forces on both raw force signals and ternary force representation by calculating the mean and variance of them during a test episode using grasping force scaling factors of 1 and 2, as shown in Table \ref{grasping_force_mean_var}. 

The result shows that different grasping forces produced significantly different signals for raw force signals but not for the ternary force representation. Since the robots trained with raw signals were not exposed to these varied force observations during training, they lack the knowledge to select appropriate actions during execution. Conversely, the ternary representation's consistency allows robots to take actions within the distribution of training samples.

\paragraph{Variations in Object Geometry} 
Force sensor data can vary significantly depending on the grasping geometry. Yet, it is common for the grasping geometry to be different, such as changes in the grasping contact point or contact area, between training and execution due to factors such as wear, tear, or deformation of the object.

To simulate different grasping geometries, we switched the manipulated object from a rectangular prism object with dimensions of 72 cm in length and a 3.6 $\times$ 3.6 cm cross-section to a cylindrical object with a length of 72 cm and a diameter of 2 cm during execution. This change resulted in an entirely different grasping geometry between training and execution. Table \ref{grasping_geometry} shows the success rate of transportation for both objects. The result indicates that, in contrast to the Raw Force method, our approach maintained a success rate of over 50\%, even if the object geometry during execution was entirely different from the one used in the training.

\begin{table}[!tp]
\centering
\caption{Success rates of transportation with different grasping geometries.}
\label{grasping_geometry}
\begin{tabularx}{\columnwidth}{>{\centering\arraybackslash}p{2.5cm} >{\centering\arraybackslash}X >{\centering\arraybackslash}X}
\toprule
\textbf{\normalsize Object Shape} &  \textbf{\normalsize Ours} & \textbf{\normalsize Raw Force} \\ 
\midrule
Rectangular prism & \textbf{82\%} & \textbf{85\%} \\ 
Cylindrical & \textbf{51\%} & 25\% \\ 
\bottomrule
\end{tabularx}
\end{table}

\section{Real Robot Experiment}
The real-robot experiment has two objectives. The first objective is to evaluate the policy's robustness against the inherent sim2real gap in grasping environment. The second objective is to evaluate the policy’s robustness when dealing with variations in object size in the real-world environment.

\subsection{Setup}
We take the best policy trained from our simulation experiments and run them directly on the real robots by building upon the physical setup used in Fig. \ref{task_fig}. GelSight Mini sensors operating at 60 Hz are used as force sensors, and OptiTrack motion capture sensors operating at 120 Hz are used to obtain the poses of the robots' end-effectors and the object. The object is made of hard cardboard.

\subsection{Result}
\subsubsection{Inherent Sim2real Gap in Grasping Environment} 
When deploying simulation-trained policy to real robots, inherent sim2real gap \cite{sim2real_3} arises due to the inability of any physics simulator to perfectly replicate real-world environments. We evaluated the robustness of our policy against this inherent sim2real gap in grasping environment, including but not limited to factors such as grasping force, force sensor models, object dynamics, and physical properties.

Despite a significant sim2real gap, including differences in grasping force and sensor properties such as noise, sensitivity, scale, and bias, robots trained using our method were still able to successfully grasp and transport the object to the target position, as shown in Fig. \ref{robot_traj}.

\subsubsection{Variations in Object Size} 
We used two different sizes of objects to simulate variations in object size. The first object was a 78.5 cm long rectangular prism object with a 4 $\times$ 4 cm cross-section; the second was 78.5 cm with a 3 $\times$ 3 cm cross-section. We assessed the robot's transportation performance by measuring the position error using the motion capture sensors, averaging the results over five experiments as shown in Table \ref{real_robot_data}. 

\begin{figure*}[tb]
    \centering
        \large
        \includegraphics[width=1.0\linewidth]{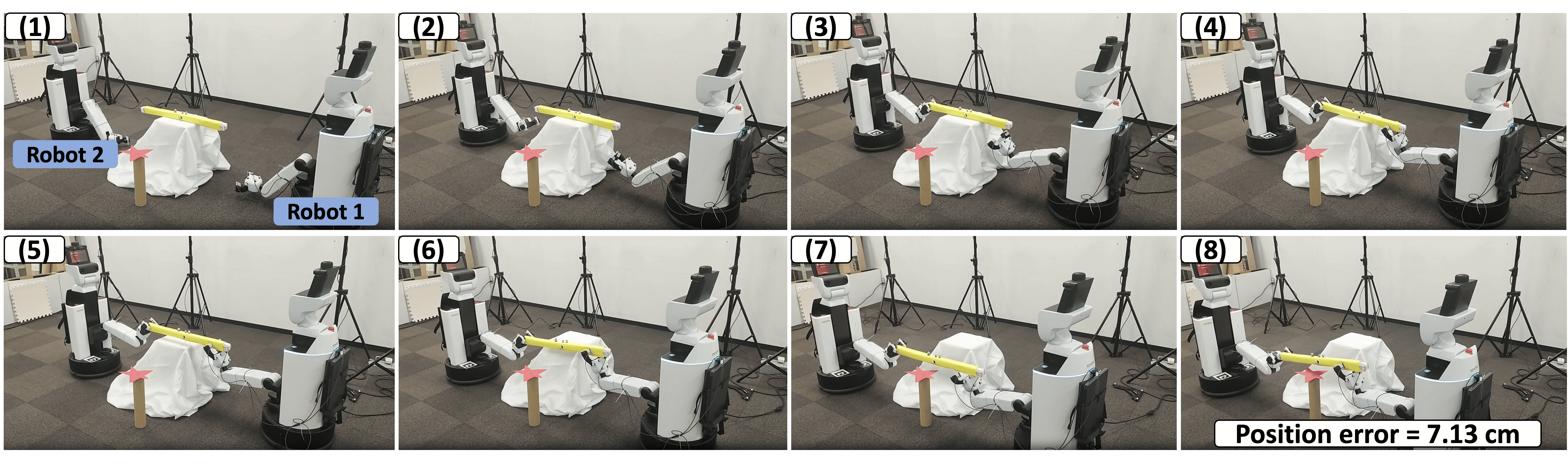}
        \caption{Successive frames of two HSRs trained by our method successfully grasp and transport an object to the target position. (1) Initially, the two robots are placed randomly around the object. (2) The robots approach their respective grasping points. (3) Robot 2 successfully grasps the object first and waits for Robot 1 to complete its grasp. (4) Robot 1 then grasps the object. The robot that first grasps the object detects its partner’s grasp by observing changes in the force signals and the orientation of the object. (5) After both robots have securely grasped the object, they lift it together. (6-7) They transport the object toward the target, coordinating through ternary force representation. (8) Finally, the object reaches the target position.}
        \label{robot_traj}
\end{figure*}

Although the position errors are big compared to the ones in the simulation, our method enables robots to transport objects despite changes in object size. Fig. \ref{robot_traj} shows the trajectories of two HSRs trained by our method successfully grasp and transport a 4 cm-sized object to the target position. On the other hand, the robots trained using the Raw Force were unable to successfully grasp the object due to a significant disparity between the force sensor model in the simulation and the real-world environment. As a result, they only hovered around the object without ever grasping it.

\begin{table}[!tp]
\centering
\caption{Object position error of our method with different object cross-section sizes. We omitted the Raw Force result as the robots were unable to successfully grasp the object.}
\label{real_robot_data}
\begin{tabularx}{\columnwidth}{>{\centering\arraybackslash}p{3.1cm} >{\centering\arraybackslash}X >{\centering\arraybackslash}X}
\toprule
\textbf{\normalsize Cross-section (cm)}  & \textbf{\normalsize Mean (cm)} & \textbf{\normalsize Variance (cm)} \\ 
\midrule
4 $\times$ 4 & 22.67 & 65.32 \\ 
3 $\times$ 3 & 30.88 & 57.19 \\ 
\bottomrule
\end{tabularx}
\end{table}



\section{Discussion}
The position error between the simulation and the real-world environment showed a significant discrepancy, revealing a notable performance gap and opportunities to improve sim2real transfer, as detailed below.

In the simulation, HSRs operated at a high frequency of 100 Hz, while in the real-world environment, they could only be operated at around 30 Hz. The robots' coordination was highly dependent on high-frequency feedback and control. In general, the higher the frequency, the more sophisticated the robots' coordination. The lower control frequency in the real-world environment caused overshooting or delays in actions, resulting in poorer coordination and higher position error.

Additionally, real HSRs experienced significant angular momentum during horizontal movements due to their omnidirectional moving mechanism, a factor not present in the simulated HSRs. This angular momentum often resulted in unintended yaw rotation, with one horizontal movement causing around 0.05 to 0.1 radian rotation. Over time, this undesired rotation accumulated, leading to misalignment of the robots' base during object transportation.

\section{Conclusion}
This paper introduces a novel MARL framework with ternary force representation to maintain consistency in force observations despite variations in the grasping environment. To address the partial observability introduced by the ternary force representation, the framework leverages an asymmetric architecture. In the real-world robot experiment, despite significant gaps between the simulation and the real environment, our method enables robots to accomplish the task.

Future work will focus on enabling the robots to autonomously learn optimal grasping points based on the number of robots involved and the geometry of the object.










\bibliographystyle{IEEEtran}
\bibliography{ref}

\begin{thebibliography}{10}
\providecommand{\url}[1]{#1}
\csname url@rmstyle\endcsname
\providecommand{\newblock}{\relax}
\providecommand{\bibinfo}[2]{#2}
\providecommand\BIBentrySTDinterwordspacing{\spaceskip=0pt\relax}
\providecommand\BIBentryALTinterwordstretchfactor{4}
\providecommand\BIBentryALTinterwordspacing{\spaceskip=\fontdimen2\font plus
\BIBentryALTinterwordstretchfactor\fontdimen3\font minus \fontdimen4\font\relax}
\providecommand\BIBforeignlanguage[2]{{%
\expandafter\ifx\csname l@#1\endcsname\relax
\typeout{** WARNING: IEEEtran.bst: No hyphenation pattern has been}%
\typeout{** loaded for the language `#1'. Using the pattern for}%
\typeout{** the default language instead.}%
\else
\language=\csname l@#1\endcsname
\fi
#2}}

\bibitem{leader_follower}
Z.~Wang and M.~Schwager, ``Kinematic multi-robot manipulation with no communication using force feedback,'' in \emph{IEEE International Conference on Robotics and Automation (ICRA)}, 2016, pp. 427--432.

\bibitem{force_implicit_2}
A.~Tsiamis, C.~K. Verginis, C.~P. Bechlioulis, and K.~J. Kyriakopoulos, ``Cooperative manipulation exploiting only implicit communication,'' in \emph{IEEE/RSJ International Conference on Intelligent Robots and Systems (IROS)}, 2015, pp. 864--869.

\bibitem{rl_dual_arm}
L.~Liu, Q.~Liu, Y.~Song, B.~Pang, X.~Yuan, and Q.~Xu, ``A collaborative control method of dual-arm robots based on deep reinforcement learning,'' \emph{Applied Sciences}, vol.~11, no.~4, 2021.

\bibitem{mappo}
C.~Yu, A.~Velu, E.~Vinitsky, J.~Gao, Y.~Wang, A.~Bayen, and Y.~Wu, ``The surprising effectiveness of {PPO} in cooperative multi-agent games,'' in \emph{Conference on Neural Information Processing Systems (NeurIPS)}, vol.~35, 2022.

\bibitem{push_2}
J.~Chen, M.~Gauci, W.~Li, A.~Kolling, and R.~Groß, ``Occlusion-based cooperative transport with a swarm of miniature mobile robots,'' \emph{IEEE Transactions on Robotics}, vol.~31, no.~2, pp. 307--321, 2015.

\bibitem{shibata}
K.~Shibata, T.~Jimbo, and T.~Matsubara, ``Deep reinforcement learning of event-triggered communication and consensus-based control for distributed cooperative transport,'' \emph{Robotics and Autonomous Systems}, vol. 159, p. 104307, 2023.

\bibitem{leader_follower_3}
M.-H. Wu, A.~Konno, and M.~Uchiyama, ``Cooperative object transportation by multiple humanoid robots,'' in \emph{IEEE/SICE International Symposium on System Integration (SII)}, 2011, pp. 779--784.

\bibitem{implicit_and_explicit}
N.~Gildert, A.~G. Millard, A.~Pomfret, and J.~Timmis, ``The need for combining implicit and explicit communication in cooperative robotic systems,'' \emph{Frontiers in Robotics and AI}, vol.~5, 2018.

\bibitem{delay_1}
P.~Grover and A.~Sahai, ``Implicit and explicit communication in decentralized control,'' in \emph{48th Annual Allerton Conference on Communication, Control, and Computing}, 2010, pp. 278--285.

\bibitem{delay_2}
V.~Villani, C.~Vercellino, and L.~Sabattini, ``How can we understand multi-robot systems? a user study to compare implicit and explicit communication modalities,'' in \emph{Distributed Autonomous Robotic Systems}, 2024, pp. 1--13.

\bibitem{marl_overview}
S.~Gu, L.~Yang, Y.~Du, G.~Chen, F.~Walter, J.~Wang, and A.~Knoll, ``A review of safe reinforcement learning: Methods, theory and applications,'' \emph{IEEE Transactions on Pattern Analysis and Machine Intelligence}, 2024.

\bibitem{marl_survey}
J.~Orr and A.~Dutta, ``Multi-agent deep reinforcement learning for multi-robot applications: A survey,'' \emph{Sensors}, vol.~23, no.~7, 2023.

\bibitem{push-and-pull}
J.~Bloom, P.~Paliwal, A.~Mukherjee, and C.~Pinciroli, ``Decentralized multi-agent reinforcement learning with global state prediction,'' in \emph{IEEE/RSJ International Conference on Intelligent Robots and Systems (IROS)}, 2023, pp. 8854--8861.

\bibitem{AAC}
L.~Pinto, M.~Andrychowicz, P.~Welinder, W.~Zaremba, and P.~Abbeel, ``Asymmetric actor critic for image-based robot learning,'' in \emph{Robotics: Science and Systems (RSS)}, 2018.

\bibitem{sim2real_3}
W.~Zhao, J.~P. Queralta, L.~Qingqing, and T.~Westerlund, ``Towards closing the sim-to-real gap in collaborative multi-robot deep reinforcement learning,'' in \emph{5th International Conference on Robotics and Automation Engineering (ICRAE)}, 2020, pp. 7--12.

\end{thebibliography}

\end{document}